\documentclass{article} 
\usepackage{iclr2025_conference,times}
\pdfoutput=1

\usepackage{amsmath,amsfonts,bm}

\def\eqref#1{equation~\ref{#1}}

\def\1{\bm{1}}

\DeclareMathAlphabet{\mathsfit}{\encodingdefault}{\sfdefault}{m}{sl}
\SetMathAlphabet{\mathsfit}{bold}{\encodingdefault}{\sfdefault}{bx}{n}

\usepackage{graphicx} 
\usepackage{hyperref}
\usepackage{url}
\usepackage{multirow}
\usepackage{float} 
\usepackage{stfloats}
\usepackage{wrapfig}   
\usepackage{booktabs}  
\usepackage{xcolor}    
\usepackage{amsmath}
\usepackage{subcaption}
\usepackage{dashrule} 
\usepackage{booktabs}
\title{FILA: Fine-Grained Vision Language Models}

\author{
\textbf{Shiding Zhu}$^1$\thanks{Equal contribution}, \textbf{Wenhui Dong}$^2$\footnotemark[1]~, \textbf{Jun Song}$^2$\footnotemark[1]\thanks{Corresponding author}, \textbf{Yingbo Wang}$^2$, \textbf{Yanan Guo}$^3$, \textbf{Bo Zheng}$^2$\\
$^1$Zhejiang University, $^2$Taobao \& Tmall Group of Alibaba, $^3$USTC
}

\iclrfinalcopy 
\begin{document}

\maketitle
\begin{abstract}
Recently, there has been growing interest in the capability of multimodal large language models (MLLMs) to process high-resolution images. A common approach currently involves dynamically cropping the original high-resolution image into smaller sub-images, which are then fed into a vision encoder that was pre-trained on lower-resolution images. While this method enables MLLMs to process high-resolution images within input constraints, it often causes semantic fragmentation by truncating objects and connected areas, leading to degraded model performance, particularly for fine-grained details and irregularly shaped objects.
To address this limitation, we propose FILA~(\textbf{F}ine-Grained V\textbf{i}sion \textbf{La}nguage Model), a novel framework designed to process images of high resolution while retaining the overall context during encoding. Specifically, we:
(i) Introduce a Hybrid Encoder that not only encodes individual sub-images but also interacts with detailed global visual features, significantly improving the model's ability to encode high-resolution images.
(ii) Propose an effective feature fusion strategy for the dynamic cropping approach, effectively leveraging information from different layers of the vision encoder.
Compared with the state-of-the-art MLLMs under the same setting, our FILA outperforms existing MLLMs in nine out of ten tasks. Specifically, FILA achieves a 9.6\% improvement in performance on the TextVQA task and a 6.9\% enhancement on the DocVQA task.
\end{abstract}    
\section{Introduction}
\label{sec:intro}
\begin{figure}[ht]
    \centering
    \begin{subfigure}{0.46\textwidth}
        \includegraphics[bb=0 0 3494 3276, width=\linewidth]{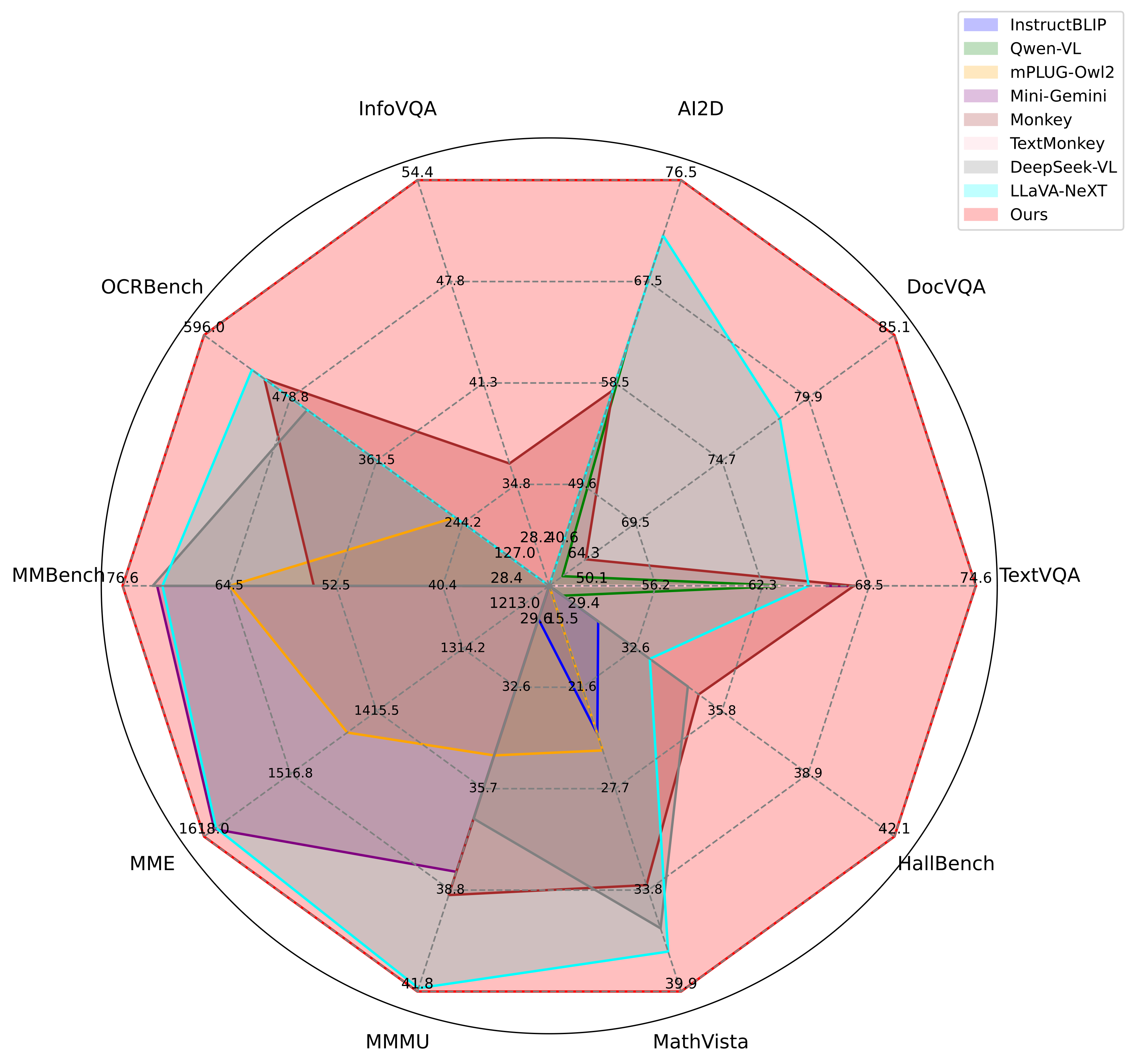}
        \caption{}
        \label{fig:image1_1}
    \end{subfigure}
    \begin{subfigure}{0.45\textwidth}
        \includegraphics[width=\linewidth]{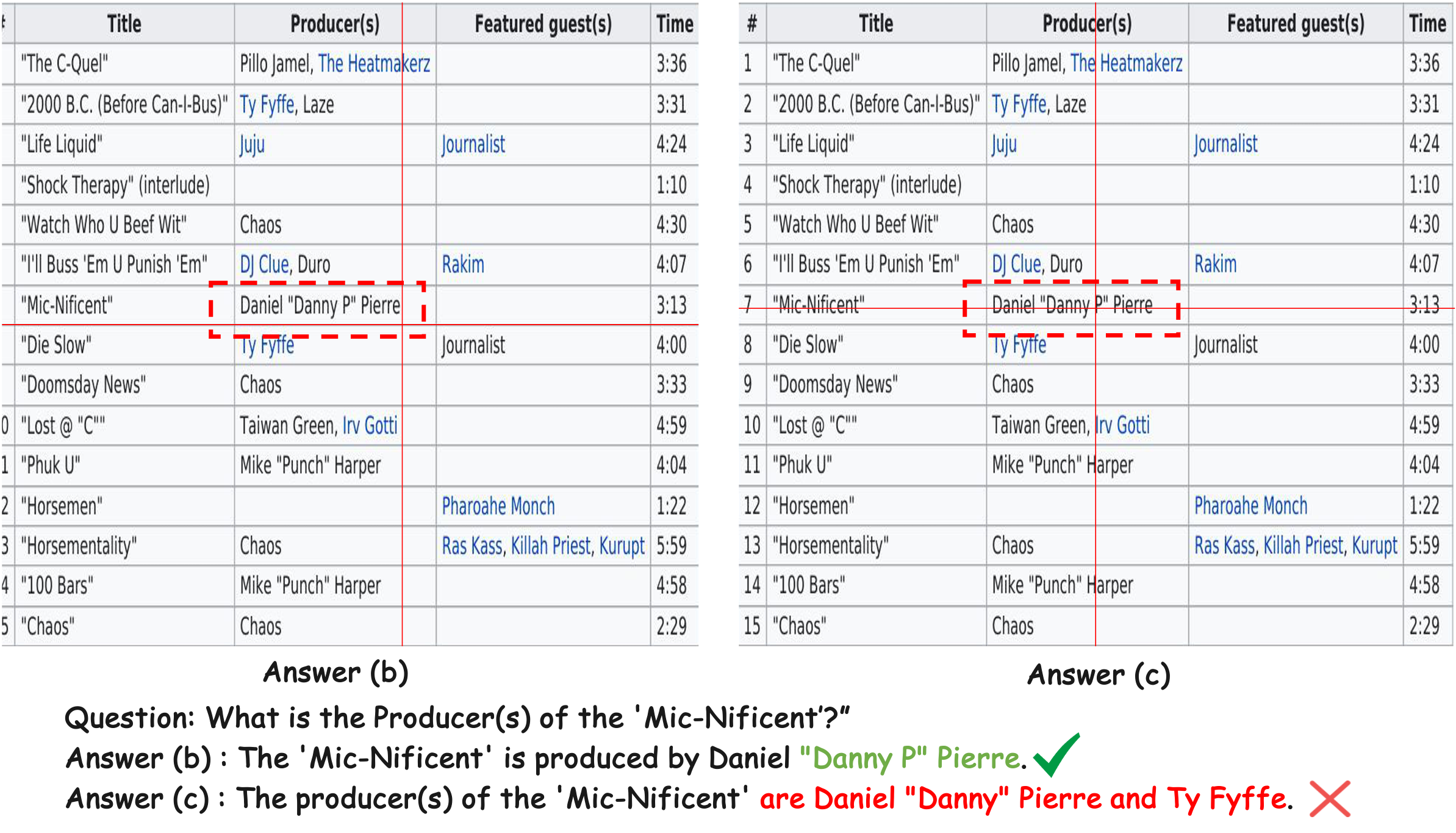} %
        \label{fig:image1_2}
    \end{subfigure}
    \caption{(a) An overall performance comparison between FILA and other existing MLLMs. (b) and (c) are examples of image fragments. The solid red line in the figures represents the cropping boundary. LLaVA-NeXT provides correct answers when pivotal information remains within the cropping boundaries. However, even slight overlaps with the boundaries cause it to misidentify the information.}
    \label{fig:image1}
\end{figure}
Inspired by the remarkable achievements of large language models (LLMs)~\citep{zhang2022optopenpretrainedtransformer,brown2020languagemodelsfewshotlearners,touvron2023llamaopenefficientfoundation}, the development of multimodal large language models (MLLMs)~\citep{alayrac2022flamingo, bai2023qwen, chen2024internvl, li2023blip,zhu2023minigpt,liu2024improvedbaselinesvisualinstruction} is advancing at a rapid pace. Researchers have directed substantial efforts towards broadening the capabilities of LLMs to encompass additional modalities, resulting in significant breakthroughs in assorted vision-language tasks. Nevertheless, the challenge of fine-grained visual recognition is not adequately addressed. This shortcoming arises partly from the constraints inherent in pre-trained visual encoders~\citep{zhai2023sigmoidlosslanguageimage,radford2021learningtransferablevisualmodels}, which often require MLLMs to process images at relatively low resolutions, typically around $224 \times 224$~\citep{zhu2023minigpt, ye2024mplug} or $336 \times 336$ ~\citep{liu2024llava}. This limitation hinders the precise identification of fine details and contextual elements within images, resulting in less accurate vision understanding and reasoning.

To solve this limitation, recent endeavors are directed toward enabling MLLMs to handle images of high resolution. One approach involves directly utilizing a visual encoder designed to process higher-resolution images. However, this approach necessitates the training of a high-quality visual encoder, which requires significant computational resources, including high-quality, high-resolution training datasets. An alternative effective strategy is dynamic cropping~\citep{ye2023ureader, li2024monkey, xu2024llava, hu2024mplug, liu2024textmonkey, liu2024llava}, where the high-resolution image is partitioned into a set of lower-resolution sub-images using adaptive algorithms.
These segmented sub-images are subsequently processed by a visual encoder that has been previously trained with lower-resolution image data, thereby augmenting the encoder's ability to capture finer details while ensuring that maintaining computational efficiency. Nevertheless, this cropping method can result in image fragmentation, thereby altering the original spatial context and disrupting positional associations within the image. This becomes particularly problematic when essential information is located at the edges of these cropped segments, as it poses challenges to encoding such information accurately. 
For instance, as depicted in Figure \ref{fig:image1} when the pivotal information does not intersect with the cropping boundaries, the model LLaVA-NeXT~\citep{liu2024llava} provides the correct answer. However, with a slight alteration in the image such that the pivotal information intersects with the cropping boundaries, LLaVA-NeXT fails to identify it correctly.

To alleviate this effect, in this paper, we propose FILA enhancing fine-grained recognition. First, we design a visual encoder, Hybrid Encoder, under the dynamic cropping strategy to address the issue of image fragmentation. Hybrid Encoder can acquire comprehensive global information to enhance the encoding of each sub-image. Furthermore, we propose a different way of visual feature interaction called ConvNeXt-ViT Deep Fusion Module (CVFM), which fully utilizes features at different levels for mutual complementarity, achieving optimal visual encoding.

\begin{figure}[htbp]
    \centering
    \begin{subfigure}{0.8\textwidth}
        \includegraphics[width=\linewidth]{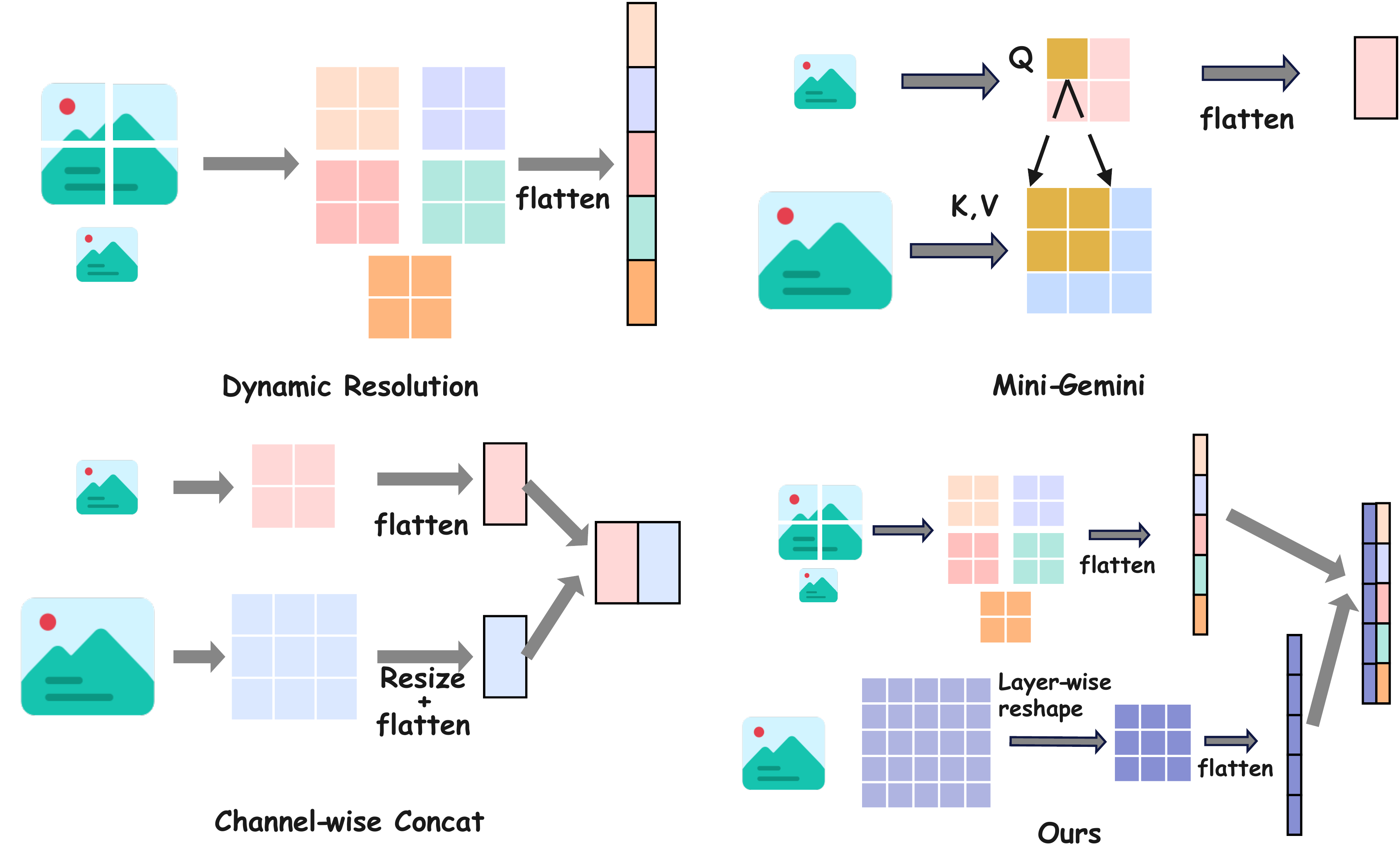} %
        \label{fig:image3}
    \end{subfigure}
    \caption{An Overview of High-Resolution Methods. This Figure illustrates the differences between our method and previous methods.}
    \label{fig:Hybrid Encoder}
\end{figure}

Figure \ref{fig:Hybrid Encoder} demonstrates the difference between our approach and previous methods. Previous methods include Dynamic Resolution, Mini-Gemini~\citep{li2024minigeminiminingpotentialmultimodality}, and Channel-wise Concat. Dynamic Resolution simply involves cropping the image and feeding it into CLIP-ViT~\citep{dosovitskiy2020image}. Mini-Gemini interacts the low-resolution CLIP-ViT features with a high-resolution auxiliary branch at the last layer, mainly using Cross Attention, while some methods utilize Channel-wise Concat. The innovation of our method lies in our interaction between two high-resolution branches and the novel idea of deep fusion, which successfully addresses the issue of image fragmentation.

To verify the effectiveness of our proposed method, we conducted extensive experiments on ten vision-language tasks, some of which include fine-grained recognition of high-resolution images, such as TextVQA~\citep{singh2019vqamodelsread} and DocVQA~\citep{mathew2021docvqadatasetvqadocument}. The experimental results show that our method outperforms existing MLLMs on 9 out of 10 Vision-Language tasks. Specifically, it surpasses LLaVA-NeXT by 9.6\% on TextVQA and by 6.9\% on DocVQA.

In summary, our contributions are three folds:
\begin{itemize}
    \item We propose a new visual encoder, Hybrid Encoder, to solve the problem of image fragmentation under dynamic cropping strategies. Hybrid Encoder can utilize comprehensive global information to enhance sub-image encoding, improving the model's upper limit in handling high-resolution images.
    \item We design a novel visual feature fusion module CVFM, cleverly utilizing information at different levels for mutual complementarity.
    \item Based on our strategy, we developed the FILA model, a powerful MLLM that outperforms existing models on multiple benchmarks, demonstrating the effectiveness of our approach.
\end{itemize}

\section{Related Work}
\label{sec:formatting}
\subsection{Large Vision-Language Model}
Driven by the tremendous success of large language models (LLMs), there has been growing interest in building end-to-end multimodal large language models. Specifically, most existing MLLMs adopt a modular structure, utilizing an intermediate network to project visual features into the word embedding space of an LLM. Then, the LLM completes various vision-language (VL) tasks in an autoregressive manner. 
The intermediate networks of existing MLLMs can generally be divided into two types: (i) learned queries, such as perceiver resamplers~\citep{alayrac2022flamingo} or Q-Formers~\citep{li2023blip}, which use fixed queries to capture features through cross-attention; and (ii) MLP modules, like those in the LLaVA series~\citep{liu2024improvedbaselinesvisualinstruction, liu2023visualinstructiontuning}. The vision encoder functions as the ``eye'' of the model, allowing it to interpret and examine visual information. This encoder can incorporate different architectures, like the Vision Transformer (ViT)~\citep{vaswani2017attention} or ConvNeXt~\citep{convnext} which have been used in image classification and object detection~\citep{zheng2023less}. A majority of MLLMs use CLIP-ViT as the vision encoder, which is trained on large image-text datasets, to proficiently extract visual features. However, CLIP-ViT is pretrained on low-resolution images and cannot directly handle high-resolution images, which limits the model's performance on fine-grained tasks.
\subsection{High-resolution MLLMS}
Directly inputting high-resolution images into visual encoders leads to high computational costs, primarily due to the quadratic complexity associated with the Transformer architecture and the substantial increase in the number of visual tokens. To alleviate this issue, existing high-resolution MLLMs can be divided into two main types. One type adopts a dynamic cropping method, cutting images into patches that are then separately input into Vision Transformers (ViTs) trained on low-resolution images. Although this method is simple and efficient, it can lead to the problem of image fragmentation, altering the original context, and disrupting positional relationships. In particular, if key information is located at the boundaries of the cropped images, it becomes difficult to encode it accurately. The other~\citep{wei2023varyscalingvisionvocabulary,zhang2024llavarenhancedvisualinstruction,hong2023cogagentvisuallanguagemodel} type uses a dual-encoder approach, introducing a high-resolution branch to supplement the low-resolution branch. For example, Vary~\citep{wei2023varyscalingvisionvocabulary} and Deepseek-VL~\citep{lu2024deepseekvlrealworldvisionlanguageunderstanding} utilize the Segment Anything Model (SAM)~\citep{kirillov2023segment} in the high-resolution visual encoder to better capture detailed information, while MiniGemini~\citep{li2024minigeminiminingpotentialmultimodality} and LLaVA-HR~\citep{luo2024feasteyesmixtureofresolutionadaptation} adopt ConvNeXt. However, the main visual encoder in these models is still CLIP-ViT, whose capabilities are limited by its low resolution.
In our work, we designed a visual encoder called Hybrid Encoder specifically for the dynamic cropping strategy. Hybrid Encoder expands the ability of CLIP-ViT to process high-resolution images, solving the problem of image fragmentation and improving the model's capacity to handle high-resolution images.

\section{Methodology}
\begin{figure*}[htbp]
    \centering
    \begin{subfigure}{0.3\textwidth}
        \includegraphics[width=\linewidth]{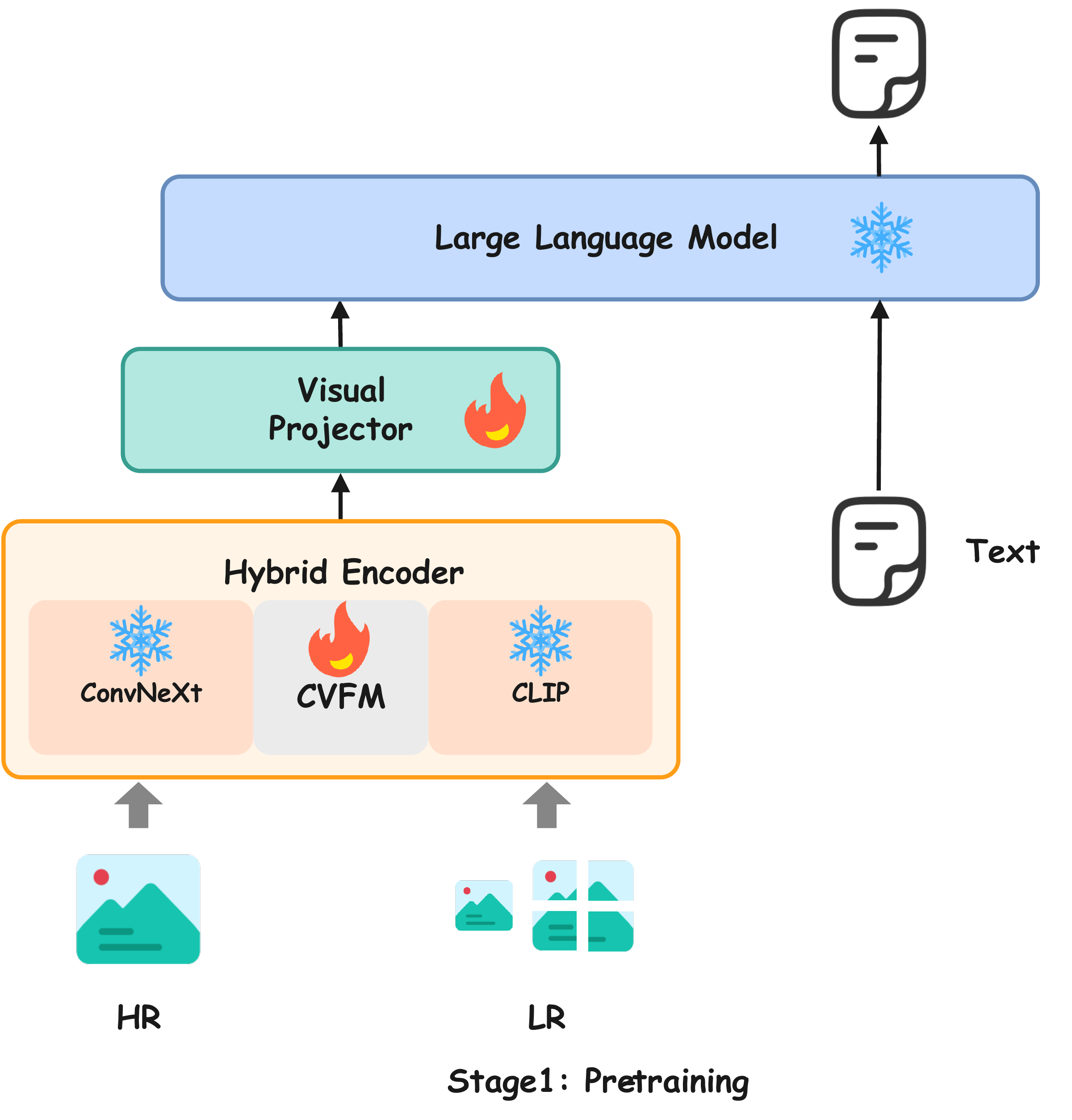} %
        \caption{}
        \label{fig:image2_1}
    \end{subfigure}
    \hfill
    \begin{subfigure}{0.3\textwidth}
        \includegraphics[width=\linewidth]{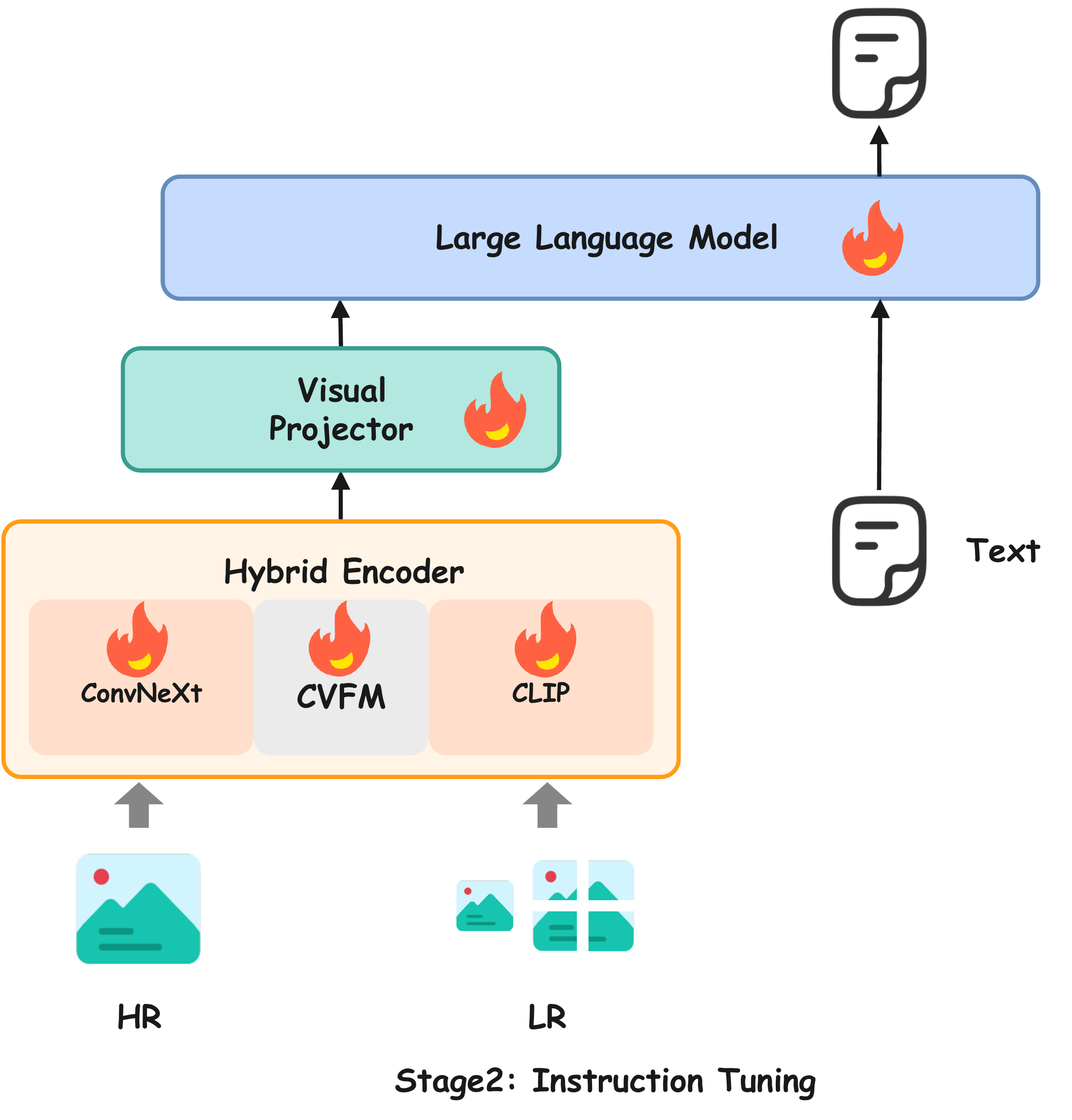} %
        \caption{}
        \label{fig:image2_2}
    \end{subfigure}
    \hfill
    \begin{subfigure}{0.3\textwidth}    
        \includegraphics[width=\linewidth]{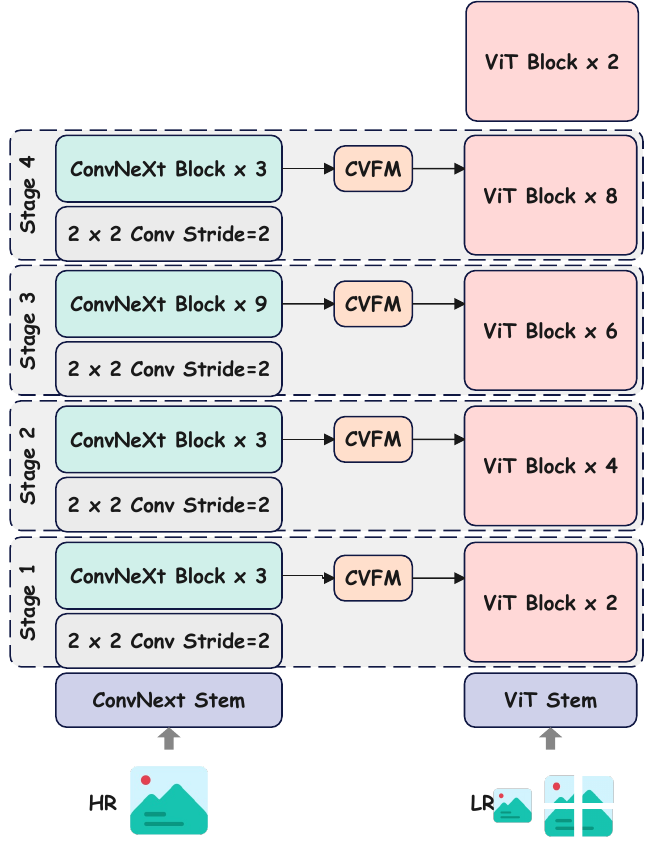} %
        \caption{}
        \label{fig:image2_3}
    \end{subfigure}
    \caption{Whole architecture for MLLMs. (a) is the first stage of training, during which we only activate the CVFM and projector. (b) is the second stage of training, where we activate all modules of the model. (c) is the structural diagram of Hybrid Encoder, which features high-resolution branches working collaboratively.}
    \label{fig:whole_architecture}
\end{figure*}
Figure~\ref{fig:whole_architecture} provides an overview of FILA. Initially, the image is dynamically segmented, resulting in several sub-images, including the original low-resolution image. These sub-images are processed by our newly designed visual encoder, the Hybrid Encoder. After passing through a projection layer, the encoded visual features are integrated with the textual features and subsequently fed into the large language model (LLM). In Section \ref{model_arch} we briefly introduce the overall model architecture. Section \ref{hybrid_encoder} presents our newly designed visual encoder, the Hybrid Encoder. In Section \ref{cvfm}, we describe our optimal fusion strategy, and Section \ref{training_para} outlines the training paradigm for the FILA model.

\subsection{Model Architecture}\label{model_arch}

We apply a dynamic segmentation method to partition the input image into smaller blocks while maintaining the original aspect ratio. The image is resized and padded to the most appropriate predefined resolution, then segmented into \(n_w \times n_h\) patches, each sized \(336 \times 336\). The predefined resolutions are: \(336 \times 672\), \(672 \times 336\), \(672 \times 672\), \(1008 \times 336\), and \(336 \times 1008\).

To select the best resolution, we calculate the effective resolution (\(Res_{eff}\)) and the wasted resolution (\(Res_{wasted}\)) for each predefined resolution. The selection aims to maximize \(Res_{eff}\) and minimize \(Res_{wasted}\) if multiple resolutions have the same \(Res_{eff}\). 
\begin{align}
\text{Scale} &= \min\left(\frac{W_h}{W_l}, \frac{H_h}{H_l}\right), \\
W_s, H_s &= W_l \times \text{Scale}, \; H_l \times \text{Scale}, \\
Res_{eff} &= \min\left(W_s \times H_s, W_l \times H_l\right), \\
Res_{wasted} &= (W_h \times H_h) - Res_{eff}.
\end{align}
Here, \(W_h\) and \(H_h\) are the predefined high resolution dimensions, while \(W_l\) and \(H_l\) are the input image's width and height. \(\text{Scale}\) is used to resize the image proportionally. After resizing, \(Res_{eff}\) ensures it does not exceed the input image's resolution, and \(Res_{wasted}\) measures unused space. The predefined resolution with the highest \(Res_{eff}\) is chosen, and if multiple resolutions have the same \(Res_{eff}\), the one with the least \(Res_{wasted}\) is selected.

In addition to the segmented patches, a global view of the image is created by resizing it to \(336 \times 336\), offering a complete overview. Both the global view and the patches are processed through a Hybrid Encoder to extract features, which are then aligned by an MLP and combined with text features. Finally, these features are input into a language decoder for the final predictions.

\subsection{Hybrid Encoder}\label{hybrid_encoder}
The architecture of the proposed Hybrid Encoder is depicted in Figure~\ref{fig:image2_3}. This architecture integrates ConvNeXt, CLIP-ViT, and CVFM modules. Both the ConvNeXt and ViT networks are divided into four stages. The final layer of each stage serves as an interaction layer, combining its output with that of the corresponding ConvNeXt stage via the CVFM module. This design enables the acquisition of fine-grained global information and effectively mitigates image fragmentation. 

In the dynamic cropping strategy, an optimal high-resolution setting \(H_h \times W_h\) is selected. The original image \(I_l\) is resized and padded to a predefined resolution while preserving its aspect ratio. As a result, a low-resolution global image and a series of sub-images \([I^{global}, I_1^{loc}, I_2^{loc}, \ldots, I_N^{loc}]\) are obtained for subsequent processing by the CLIP-ViT. 
In parallel, the original image \(I_l\) is also resized and padded to a higher resolution of $(\frac{32}{14} \times H_h, \frac{32}{14} \times W_h)$. This high-resolution image \(I_h\) is directly encoded by ConvNeXt, yielding a set of multi-level high-resolution features \(\mathbf{F}_{vh} = [\mathbf{F}_{vh}^{1}, \mathbf{F}_{vh}^{2}, \mathbf{F}_{vh}^{3}, \mathbf{F}_{vh}^{4}]\).

The choice of the $(\frac{32}{14} \times H_h, \frac{32}{14} \times W_h)$ resolution ensures that the image aspect ratio is maintained, allowing more detailed information to be captured. It also facilitates subsequent alignment with CLIP-ViT features. In particular, this configuration guarantees that the output dimensions of each ConvNeXt stage are integer multiples of the concatenated dimensions of all ViT hidden layer features, simplifying the alignment and dimensional transformations. Detailed dimensional operations are provided in Table~\ref{tab:Alignment Strategy}. In this table, we use an input image of \(336 \times 336\) as an example, and the cropping strategy dictates that the resolution of the image input to ConvNeXt is \(768 \times 768\).

During the encoding process in CLIP-ViT, the features of each sub-image interacts with a corresponding subset of \(\mathbf{F}_{vh}\), thus incorporating enhanced contextual information and effectively addressing the fragmentation problem associated with conventional dynamic slicing methods. The specific mechanisms of these interactions are elaborated upon in the following section.



\subsection{ConvNeXt-ViT Deep Fusion Module (CVFM)}\label{cvfm}
To enhance the interaction between ConvNeXt and CLIP-ViT features, we propose the ConvNeXt-ViT Deep Fusion Module (CVFM). The CVFM is designed to crop the overall features extracted by ConvNeXt to match the corresponding features of CLIP-ViT, concatenate them along the channel dimension, and integrate features from various stages of ConvNeXt into the encoding process of CLIP-ViT.

Initially, the features from each stage of ConvNeXt are resized to align with the feature dimensions of the hidden layers in CLIP-ViT. Consistent with the outcomes of dynamic cropping, the features from each stage, denoted as $\mathbf{F}_{vh}^{i}$, are segmented into global and multiple local components:\[
\mathbf{F}_{vh}^{i} = \left[\mathbf{F}_{vh-i}^{\text{global}}, \mathbf{F}_{vh-i}^{\text{loc1}}, \mathbf{F}_{vh-i}^{\text{loc2}}, \ldots, \mathbf{F}_{vh-i}^{\text{locN}}\right]\]
In the interaction layer, the hidden layer features of CLIP-ViT are fused with the corresponding high-resolution features. Specifically, lower-level features interact with lower-level features, and higher-level features interact with higher-level features. The interaction is formalized as follows:
\begin{align}
\mathbf{F}_{vl}^{i'} = \mathbf{F}_{vl}^i + \tanh\left(\alpha_{\text{dense}}\right) \cdot \text{MLP} \left( \mathbf{F}_{vl}^i \oplus \mathbf{F}_{vh}^{i'} \right),
\end{align}
where $\mathbf{F}_{vl}^i$ represents the feature before the interaction layer of the $i$-th stage of CLIP-ViT, and $\mathbf{F}_{vh}^{i'}$ denotes the corresponding high-resolution feature. The $\oplus$ operator signifies concatenation along the channel dimension. After passing through the multilayer perceptron (MLP), the number of channels is restored to match the original number of channels in CLIP-ViT. The parameter $\alpha_{\text{dense}}$ serves as a gating mechanism, initialized to zero to preserve the integrity of the visual model during initialization, thereby enhancing both stability and performance.

Although each CLIP-ViT interacts exclusively with its corresponding high-resolution feature, the convolutional properties of ConvNeXt enable each feature to effectively capture global information, thereby compensating for the limitation of each ViT, which is restricted to local sub-images. Additionally, the interaction layer provides ViT with more detailed features, resulting in a final visual representation that is both more accurate and comprehensive.

\subsection{Training Paradigm}\label{training_para}
As shown in Figure \ref{fig:whole_architecture}, we adopted a two-stage approach in training FILA, which includes low-resolution pre-training and high-resolution visual instruction fine-tuning. The goal is to align visual features with the language model during the pre-training stage, followed by fine-tuning the language model in the instruction tuning stage. Specifically, during the pre-training phase, we do not use a dynamic cropping strategy; instead, we directly resize the images to 336 $\times$ 336 for \( I_l \) and 768 $\times$ 768 for \( I_h \). We enable CVFM and the visual projector to train while keeping the other layers of the language model and Hybrid Encoder frozen. In the instruction fine-tuning stage, we employ a dynamic cropping strategy, and all parts of the model are trainable, including the ConvNeXt used for generating high-resolution features. The detail training setting as described in Section \ref{sec:training-settings}. 

\section{Experiments}
\subsection{Implementation}
\setlength{\tabcolsep}{3pt} 
\renewcommand{\arraystretch}{1.2} 


\textbf{Datasets} To achieve effective cross-modality alignment and instruction finetuning, we have gathered high-quality datasets from publicly accessible sources. Initially, we focus on pretraining the projector and interaction layers by amassing approximately 1.2M image captions. This includes 558K image-caption pairs extracted from the LLaVA-filtered CC3M dataset ~\citep{changpinyo2021conceptual12mpushingwebscale} and 695K GPT-4V-generated captions from the ALLaVA dataset~\citep{chen2024allavaharnessinggpt4vsynthesizeddata}. For instruction finetuning, diverse datasets are utilized: 643K single- and multi-turn conversations (minus 21K TextCaps data) from the LLaVA~\citep{liu2023visualinstructiontuning} dataset, in addition to 100K QA pairs from ShareGPT4V~\citep{chen2023sharegpt4vimprovinglargemultimodal}, 10K LAION-GPT-4V captions~\citep{gpt4vdataset}, 6K text-only multi-turn dialogues from LIMA~\citep{zhou2023limaalignment} and OpenAssistant2~\citep{kopf2024openassistant}, and 700K GPT-4V-constructed instruction pairs from the ALLaVA dataset. To enhance OCR-related competencies, we further incorporate 28K QA pairs comprised of 10K DocVQA~\citep{mathew2021docvqadatasetvqadocument}, 4K ChartQA~\citep{masry2022chartqabenchmarkquestionanswering}, 10K DVQA~\citep{kafle2018dvqaunderstandingdatavisualizations}, and 4K AI2D~\citep{ai2d} data. Overall, our datasets support about 1.5M instruction-related conversations aimed at improving image comprehension.

\subsection{Evaluations}
\textbf{Benchmarks} Following prior work, we evaluated our HyViLM on ten benchmarks. These include five general multimodal understanding benchmarks: MMBench~\citep{mmbench}, MMMU~\citep{yue2023mmmu}, MME~\citep{mme}, MathVista~\citep{lu2024mathvista}, and HallusionBench~\citep{guan2024hallusionbenchadvanceddiagnosticsuite}, as well as five document understanding-related benchmarks: TextVQA~\citep{singh2019vqamodelsread}, DocVQA~\citep{docvqa}, AI2D~\citep{ai2d}, InfoVQA~\citep{mathew2021infographicvqa}, and OCRBench~\citep{liu2024ocrbenchhiddenmysteryocr}.
This selection was made to demonstrate that our model not only possesses strong fine-grained recognition abilities but also exhibits excellent general capabilities.

\noindent \textbf{MLLMs} 
We compared our model with some state-of-the-art multi-language large models (MLLMs). (1) Normal Resolution MLLMs, such as LLaVA1.5~\citep{liu2024improvedbaselinesvisualinstruction}, InstructBLIP~\citep{instructblip}, Qwen-VL~\citep{bai2023qwen}, and mPLUG-Owl2~\citep{ye2024mplug}. (2) High Resolution MLLMs, such as MiniGemini~\citep{li2024minigeminiminingpotentialmultimodality}, CogAgent~\citep{hong2023cogagentvisuallanguagemodel}, Monkey~\citep{li2024monkey}, TextMonkey~\citep{liu2024textmonkey}, HiRes-LLaVA-336px~\citep{huang2024hiresllavarestoringfragmentationinput}, LLaVA-UHD~\citep{xu2024llava}, DocOwl-1.5-Chat~\citep{hu2024mplug}, DeepSeek-VL~\citep{lu2024deepseekvlrealworldvisionlanguageunderstanding}, MiniGemini-HD~\citep{li2024minigeminiminingpotentialmultimodality}, and LLaVA-NeXT~\citep{liu2024llava}.

\subsection{Main Results}

\begin{table*}[htbp]
  \centering
  \resizebox{\textwidth}{!}{ 
  \begin{tabular}{ccc|ccccc|ccccc}
    \toprule
    \multirow{2}{*}{\textbf{Model}} & \multirow{2}{*}{\textbf{LLM}} & \multirow{2}{*}{\textbf{MaxRes}} & \multicolumn{5}{c|}{\textbf{Doc}} & \multicolumn{5}{c}{\textbf{General}} \\
    & & & TextVQA & DocVQA & AI2D & InfoVQA & OCRBench & MMBench & MME & MMMU & MathVista & HallBench \\
    \midrule
    \multicolumn{13}{c}{\textbf{Normal Resolution MLLMs}} \\
    \midrule
    LLaVA1.5 & Vicuna-7B & $336\times336$ & 58.2 & - & 54.8 & - & - & 64.3 & 1511/- & - & - & - \\
    InstructBLIP & Vicuna-7B & $224\times224$ & 50.1 & - & 40.6* & - & 276* & 28.4* & 1213/- & 30.6* & 24.4* & 31.2* \\
    Qwen-VL & Qwen-7B & $448\times448$ & 63.8 & 65.1 & 62.3 & - & 127* & 38.2 & - & 29.6* & 15.5* & 29.9* \\
    mPLUG-Owl2 & LLaMA-7B & $224\times224$ & 58.2 & - & 55.7* & - & 255* & 64.5 & 1450/- & 34.7* & 25.4* & 29.4* \\
    \midrule
    \multicolumn{13}{c}{\textbf{High Resolution MLLMs}} \\
    \midrule
    MiniGemini & LLaMA3-8B & $336(768)$ & 67.2 & - & - & - & - & 72.7 & 1606/341 & 38.2 & - & - \\
    CogAgent & LLaMA2-7B & $224(1120)$ & - & 81.6 & - & 44.5 & - & 76.1 & - & - & - & - \\
    Monkey & Qwen-7B & $1344\times896$ & 67.6 & 66.5 & 57.9 & 36.1 & 514* & 55.0* & - & 38.9* & 33.5* & 34.9* \\
    TextMonkey & Qwen-7B & $896$ & 65.9 & 64.3 & - & 28.2 & 561 & - & - & - & - & - \\
    HiRes-LLaVA-336px & Vicuna-7B & $1344\times1344$ & - & 74.7 & 69.7 & 48.0 & - & 70.5 & - & - & - & \textbf{42.6} \\
    LLaVA-UHD & Vicuna-13B & $672\times1008$ & - & - & - & - & - & 68 & 1535/ & - & - & - \\
    DocOwl-1.5-Chat & 7B & $1344\times1344$ & 68.6 & 82.2 & - & 50.7 & - & - & - & - & - & - \\
    DeepSeek-VL & DeepSeek-LLM-7B & 384(1024) & - & - & 65.3* & - & 456 & 73.2 & - & 36.6 & 36.1 & 34.5* \\
    MiniGemini-HD & LLaMA3-8B & $672(1536)$ & 71.6 & - & - & - & - & - & 1532/357 & 37.0 & - & - \\
    LLaVA-NeXT & LLaMA3-8B & $672\times672$ & 65.0 & 78.2 & 71.6 & - & 531* & 72.1 & 1604/368 & 41.7 & 37.5 & 33.1* \\
    \midrule
    HyViLM & LLaMA3-8B & 672(1536) & \textbf{74.6} & \textbf{85.1} & \textbf{76.5} & \textbf{54.4} & \textbf{596} & \textbf{76.6} & \textbf{1618/388} & \textbf{41.8} & \textbf{39.9} & 42.1 \\
    \bottomrule
  \end{tabular}
}
  \caption{Quantitative results on 10 popular benchmarks, including 5 general benchmarks and 5 document understanding-related benchmarks. ``MaxRes'' means the maximum
resolution supported. The meaning of ``672(1536)'' is to integrate features of an image with a resolution of $1536\times1536$ into features of an image with a resolution of $672\times672$. * means that the metric is provided by OpenCompass~\citep{2023opencompass}.}
  \label{tab:1}
\end{table*}

\textbf{General Multimodal Understanding} 
We evaluated HyViLM in terms of general multimodal understanding, and the results are shown in Table \ref{tab:1}. Our model exceeds the vast majority of models of comparable size. In particular, on MME, we surpass LLaVA-NeXT by 1.7\% and MiniGemini-HD by 6.2\%. MiniGemini-HD also employs the ConvNeXt visual encoder, with a maximum image resolution consistent with ours, and the training data are almost identical to ours, highlighting the superiority of our approach compared to Minigemini. On MathVista, we significantly outperform the best model, LLaVA-NeXT, by 6.4\%. Furthermore, on three additional general benchmarks, MMBench, MMMU and HallBench, we also achieved SOTA or near-SOTA results. It is evident that our approach greatly enhances the general understanding capabilities of MLLMs.

\noindent \textbf{Document Understanding}
We have achieved outstanding performance on document-related leaderboards, making significant advancements in five document understanding tasks. Compared to the state-of-the-art models, HyViLM outperforms by 4.2\% on TextVQA, 3.5\% on DocVQA, 6.8\% on AI2D, 7.3\% on InfoVQA, and 6.2\% on OCRBench. Additionally, our input tokens to the LLM remain consistent with LLaVA-NeXT, resulting in a substantial performance boost while maintaining computational complexity. It is noteworthy that MiniGemini-HD is a variant of our model that only enables shallow interactions of different visual features, highlighting the importance of deep visual feature interactions under a dynamic cropping strategy. Therefore, we can conclude that our designed Hybrid Encoder significantly enhances the model’s fine-grained recognition capability.

\subsection{Ablation studies}
In this section, we conduct ablation experiments on our model to fully validate its effectiveness. We use TextVQA, InfoVQA, MME for the ablation study.
\begin{table}
  \centering
  \small
  \begin{tabular}{c|ccc}
    \toprule
    Method & TextVQA & InfoVQA & MME \\
    \midrule
    DS & 71.4 & 46.7 & 2000 \\
    DS + Channle Concat & 71.0 & 50.0 & 1989 \\
    DS + Local Cross Attetion & 73.2 & 50.6 & 2043\\
    DS + Global Cross Attetion & 71.5 & 47.4 & 2000\\
    DS + Hybrid Encoder & 74.6 & 54.4 & 2006\\
    \bottomrule
  \end{tabular}
  \caption{The Impact of Image Fragmentation on the Model. Except for Hybrid Encoder, the other feature interactions in the table occur in the final layer of the ViT. ``DS'' stands for dynamic slicing.}
  \label{tab:study_1}
\end{table}

\noindent \textbf{Impact of Image Fragmentation} To validate our point that the dynamic slicing strategy can lead to image fragmentation, we compared our method with several alternatives: simply using dynamic slicing to improve model resolution, and a variant of our model that combines dynamic slicing with feature interaction between CLIP-ViT and ConvNeXt at the last layer. Interaction methods include channel-wise concatenation, local cross-attention, and global cross-attention. Local cross-attention means that each feature of CLIP-ViT only interacts with its corresponding feature from ConvNeXt, whereas global cross-attention means that each feature of CLIP-ViT interacts with all features from ConvNeXt. We trained using the same complete dataset. As shown in Table \ref{tab:study_1}, our model significantly outperforms the dynamic slicing approach. Compared to dynamic slicing with last-layer interaction, our model also shows a noticeable advantage, which rules out the impact of ConvNeXt on model performance. It also indicates that the issue of image fragmentation occurs during the encoding process of CLIP-ViT, and that performing feature interaction after encoding offers limited performance improvement. These experiments clearly demonstrate that it's necessary to alter the internal structure of CLIP-ViT, so it becomes an integrated whole capable of encoding high-resolution images holistically.

\begin{table}
  \centering
  \small
  \begin{tabular}{c|c|ccc}
    \toprule
    Conv Layer & ViT Layer & TextVQA & InfoVQA & MME \\
    \midrule
    Multi Layers & 2 & 73.5 & 54.02 & 2001 \\
    Multi Layers & 6 & 73.9 & 53.3 & 2020 \\
    Last Layer & 4 & 73.7 & 53.4 & 2019 \\
    Pyramid Fusion & 4 & 73.7 & 53.6 & 1993 \\
    Multi Layers & 4 & 74.6 & 54.4 & 2006 \\
    \bottomrule
  \end{tabular}
\caption{Ablation study of the optimal interaction structure. "Multi Layers" refers to the interaction between lower- and higher-level features of ConvNeXt and ViT. "Last Layer" uses only the final layer features of ConvNeXt for interaction. "Pyramid Fusion" fuses multi-level features of ConvNeXt before interacting with ViT. "ViT Layer" indicates the number of interaction layers in ViT.}
  \label{tab:study_2}
\end{table}

\noindent \textbf{The Optimal Interaction Structure}
In Table \ref{tab:study_2}, we conducted ablation experiments to explore the optimal interaction structure. As shown in Table \ref{tab:study_1}, using only the output from the last layer of ConvNeXt for interaction features provides limited additional information and proves to be suboptimal. Therefore, we designed a multi-layer to multi-layer interaction scheme, where the lower-layer features of ConvNeXt interact with the shallow-layer features of CLIP, and the deep-layer features of ConvNeXt interact with the deep-layer features of CLIP. We believe that this approach allows the CLIP model to gradually integrate into the ConvNeXt features while maintaining feature alignment. We also explored a pyramid-like approach, where we first fused multiple layers of features from ConvNeXt before interacting with CLIP-ViT, but the results were not satisfactory. On top of this, we also varied the number of interaction layers and found that four layers are optimal. The performance of the model declines with two or six layers. We believe that too few interaction layers cannot solve the issue of image fragmentation, while too many layers disrupt the excellent encoding capabilities of the original CLIP-ViT.

\begin{table}
  \centering
  \small
  \begin{tabular}{c|c|ccc}
    \toprule
    Resize Method & Int. Mode & TextVQA & InfoVQA & MME \\
    \midrule
    Interp & Local CA & 72.8 & 50.0 & 1980 \\
    Interp & Global CA & 71.3 & 49.0 & 1997\\
    Interp & Add & 73.9 & 53.3 & 2045 \\
    Conv & Channel & 74.4 & 54.9 & 1973 \\
    Interp & Channel & 74.6 & 54.4 & 2006 \\
    \bottomrule
  \end{tabular}
  \caption{Ablation Study of Interaction Methods. \textbf{Int. Mode} means the interaction mode between ViT and ConvNeXt in CVMF, and \textbf{CA} refers to Cross Attention. Note that the experiments here are all based on deep feature interactions.}
  \label{tab:study 3}
\end{table}

\noindent \textbf{Ablation Study of Interaction Methods}
The above ablation study demonstrates that utilizing multi-level information from ConvNeXt to interact with CLIP-ViT yields the best results. The following ablation experiments investigate different fusion methods. We designed the experiments as follows: in the CVFM, we used channel concatenation, local cross attention, global cross attention, and feature addition as fusion methods. It can be seen that the channel concatenation method produced the best results. This method requires resizing the features from different levels of ConvNeXt to match the feature dimensions of CLIP-ViT. We compared interpolation and convolution as resizing methods and found that interpolation achieves slightly better performance than convolution.

\subsection{Qualitative Analysis}

\begin{figure*}[htbp]
    \centering
    \includegraphics[width=0.8\textwidth]{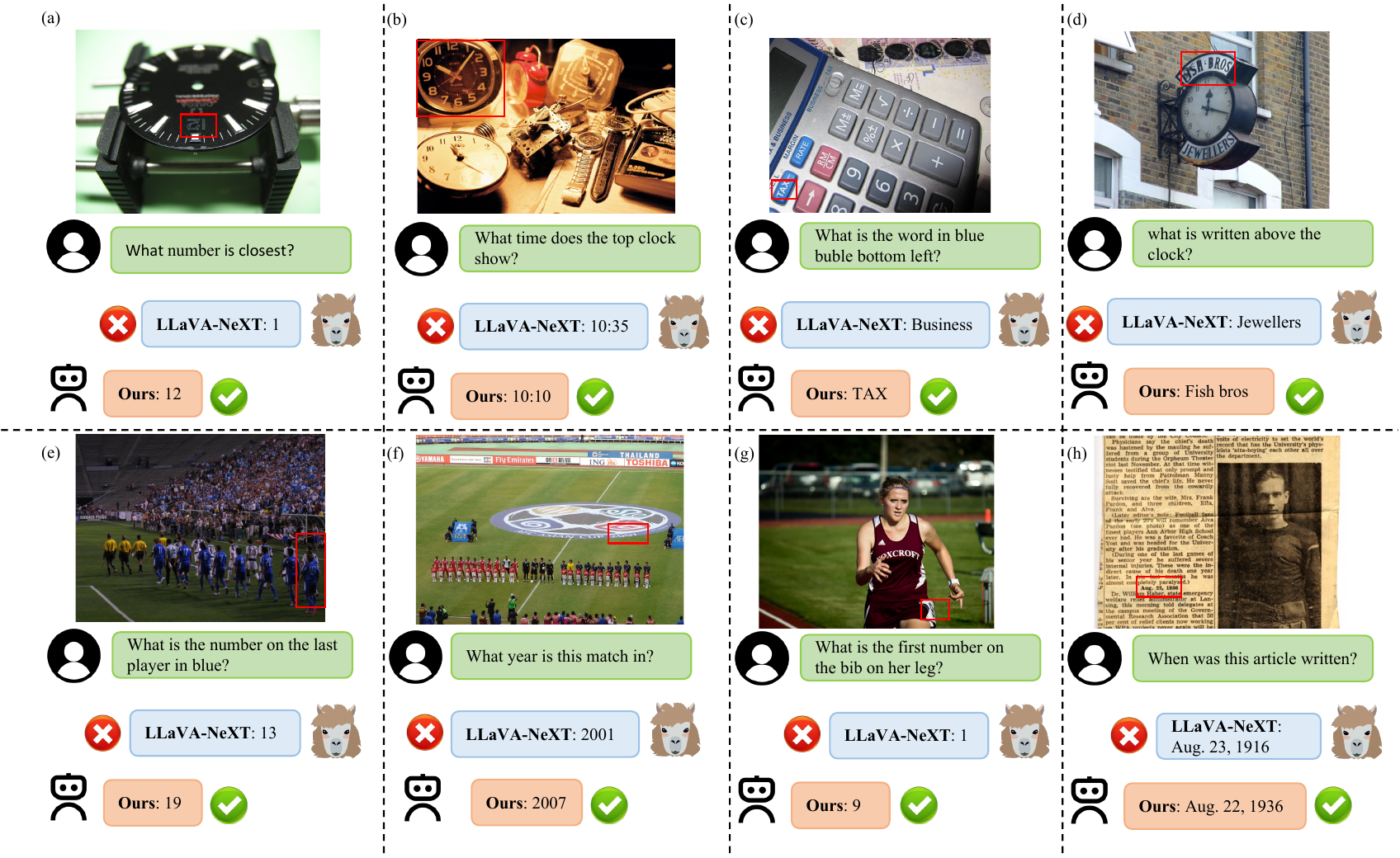} 
    \caption{Qualitative Analysis. The issue of image fragmentation has been greatly reduced after incorporating Hybrid Encoder.} 
    \label{fig:case2} 
\end{figure*}

In Figure \ref{fig:case2}, we compare the prediction results of HyViLM with LLaVA-NeXT, where LLaVA-NeXT is trained with the same data. Our model has made advancements on three levels: recognition of critical information near the cropped boundaries, judgment of positional relationships, and detail recognition capability. In the example (a), the correct answer ``12'' was split into two sub-images, causing LLaVA-NeXT to make a mistake and only answer ``1'', while our model answered correctly. In the example (d), due to dynamic image cropping, LLaVA-NeXT misjudged the positional relationship of top and bottom, as the question required the text on the clock to be answered, but it responded with the text below the clock. In the example (h), our model accurately recognized very fine-grained information and was entirely correct, whereas LLaVA-NeXT only partially succeeded in recognizing the information. These results further validate the effectiveness of Hybrid Encoder in addressing image fragmentation and improving the model's fine-grained recognition capability.
\section{Conclution}
In this article, we introduce a novel visual encoder,  Hybrid Encoder, under a dynamic partitioning strategy.  Hybrid Encoder addresses the fragmentation issue in images effectively by altering the structure of CLIP-ViT and incorporating global and fine-grained features using the CVMF approach. The model HyViLM, which applies  Hybrid Encoder, achieves state-of-the-art performance in extensive experiments, including both document-related benchmarks and various general benchmarks. Notably, HyViLM is the first study to explore the collaboration of two high-resolution branches, offering a new perspective for further enhancing the fine-grained recognition capabilities of MLLMs.

\bibliography{main.bbl}
\bibliographystyle{iclr2025_conference}

\appendix
\clearpage
\setcounter{page}{1}

\section{More Visualization}
\begin{figure*}[htbp]
    \centering
    \includegraphics[width=\textwidth]{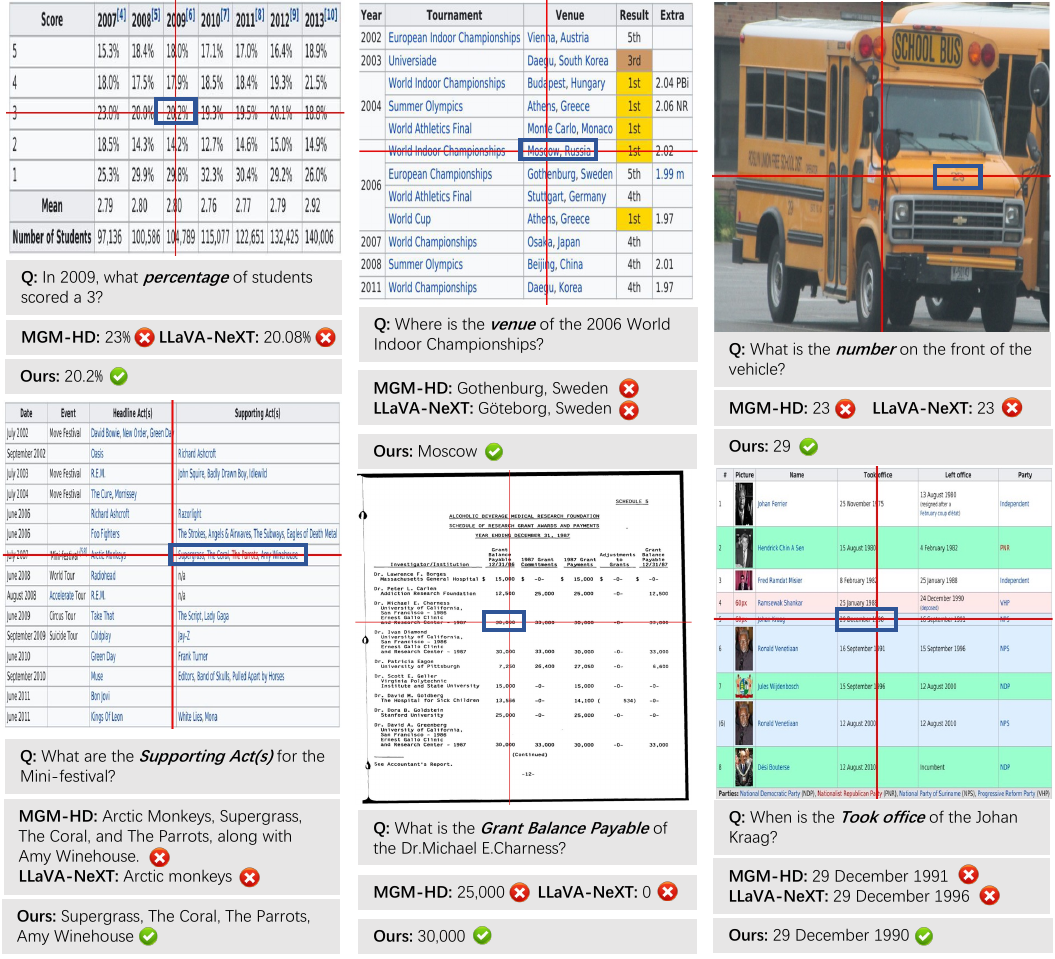}
    \caption{Qualitative results. The red solid line in the image represents the cropping boundary, and the blue box indicates the location of the answer. All images are sized at \( 672 \times 672 \). Neither the red solid line nor the blue box will be input into the model.}
    \label{fig:supple}
\end{figure*}

We selected images from DocVQA, WikiTableQuestions, and TextVQA, designed questions, and shifted the images so that the answers are at the edge of the cropped boundary. We compared our model with Minigemini-HD and LLaVA-NeXT. As shown in Figure \ref{fig:supple}, our model's performance in answering at the cropped boundary is mostly correct, while both Minigemini-HD and LLaVA-NeXT provided incorrect answers.

\section{Training Settings}\label{sec:training-settings}
In this study, we use the ViT-L pretrained by CLIP~\citep{radford2021learningtransferablevisualmodels} and ConvNeXt-L pretrained by LAION~\citep{schuhmann2022laion5bopenlargescaledataset} as the high-resolution visual encoder and the main architecture of Hybrid Encoder, respectively. For the language model, we utilize LLaMA3-8B-Instruct~\citep{touvron2023llamaopenefficientfoundation}. We optimize the model for 1 epoch using the AdamW optimizer with a cosine learning rate schedule and a warmup ratio of 0.03. During the pretraining stage, we use a peak learning rate of 1e-3 and a batch size of 256, training the projection and interaction layers. In the instruction fine-tuning phase, our main peak learning rate is 1e-5 with a batch size of 64. The learning rate for the visual encoder is reduced to 2e-6 to ensure stability, while the interaction layer remains at 1e-5. We employ the DeepSpeed Zero~\citep{rajbhandari2020zeromemoryoptimizationstraining} 2 strategy, completing the optimization in a total of 32 hours on 32 $\times$ A800 GPUs.

\section{Alignment Strategy}

\begin{table*}[htbp]
  \centering
  \resizebox{\textwidth}{!}{
  \begin{tabular}{c|c|c|c|c}
    \toprule
    Conv Stage & Input Dimensions (D, H, W) & Output Dimensions (D, H, W) & ViT Layer & ViT Dimensions (D, H, W) \\
    \midrule
    1 & (192, 192, 192) & (3072, 24, 24) & 2 & (1024, 24, 24) \\
    2 & (384, 96, 96) & (1536, 24, 24) & 6 & (1024, 24, 24)\\
    3 & (768, 48, 48) & (3072, 24, 24)& 12 & (1024, 24, 24)\\
    4 & (1536, 24, 24) & (1536, 24, 24) & 20 & (1024, 24, 24)\\
    \bottomrule
  \end{tabular}
  }
  \caption{Alignment Strategy. The alignment strategy ultimately aims to ensure that ConvNeXT aligns with the hidden states of CLIP-ViT along the H and W dimensions, allowing concatenation along the channel dimension.}
  \label{tab:Alignment Strategy}
\end{table*}

In the ConvNeXT-ViT Deep Fusion Module, we opted for channel concatenation. Table \ref{tab:Alignment Strategy} provides a detailed illustration of the dimensional changes in the features at different stages of ConvNeXT and the corresponding interaction layers with CLIP-ViT. Here, we use an image with a resolution of \( 372 \times 372 \) as an example. According to the cropping strategy, the resolution of the images input into ConvNeXT will be \( 768 \times 768 \).

\end{document}